\documentclass{article}
\usepackage[utf8]{inputenc}
\usepackage{amsmath}
\usepackage{amsfonts}
\usepackage{amssymb}
\usepackage{graphicx}
\usepackage{algorithm}
\usepackage{algpseudocode}
\usepackage{tikz}
\usetikzlibrary{positioning, arrows.meta, calc} 
\usepackage{fontenc} 
\usepackage{booktabs} 
\usepackage{tabularx} 
\usepackage{array} 
\usepackage{float} 
\usepackage{cite}

\title{Dynamic Policy Induction for Adaptive Prompt Optimization: Bridging the Efficiency-Accuracy Gap via Lightweight Reinforcement Learning}
\author{\small Jiexi Xu \\ \small University of California, Irvine \\ \small School of Information \& Computer Science}
\date{\small \today}

\begin{document}

\maketitle

\begin{abstract}
The efficacy of Large Language Models (LLMs) on complex reasoning tasks is highly dependent on the employed prompting strategy.
Static methodologies, such as relying universally on Chain-of-Thought (CoT) or Few-Shot prompting, impose a rigid trade-off between performance maximization and resource consumption \cite{brown2020language}.
Highly accurate but computationally expensive strategies, like Self-Consistency (SC), result in significant operational waste when applied to simple tasks, while inexpensive strategies often fail on complex inputs.
This paper formalizes the selection of an optimal prompting strategy as a single-step Markov Decision Process (MDP) and proposes the \textbf{Prompt Policy Network (PPN)}.
The PPN, a lightweight neural network external to the generative LLM, learns an optimal policy $\pi(a|s)$ using Proximal Policy Optimization (PPO).
Optimization is driven by a resource-explicit composite reward function, $R = \alpha \cdot \text{Accuracy} - \beta \cdot \text{Computational Cost}$.
Empirical evaluation on arithmetic reasoning benchmarks demonstrates that the PPN dynamically allocates high-cost strategies only when necessary, achieving superior performance on the efficiency-accuracy Pareto front compared to fixed-strategy and heuristic baselines, yielding substantial computational savings (e.g., 61.5\% token cost reduction) while maintaining competitive accuracy.
This work contributes a systematic, adaptive framework necessary for efficient and cost-effective deployment of LLMs, aligning with contemporary efforts in lightweight language model optimization \cite{chen2024adaptive}.
\end{abstract}


\section{Introduction}

\subsection{Motivation: The Static Prompting Bottleneck and Efficiency Crisis}

The widespread deployment of Large Language Models (LLMs) across diverse tasks has underscored the importance of effective Prompt Engineering.
While advanced strategies like Chain-of-Thought (CoT) prompting \cite{wei2022chain} and Self-Consistency (SC) \cite{wang2022self} are critical for robust reasoning, their generalized application across heterogeneous inputs imposes a rigid trade-off between performance and resource consumption \cite{brown2020language,li2025ammkd}.
The core issue lies in the computational overhead of complex prompts.
Highly effective reasoning strategies, such as SC, necessitate multiple parallel inference calls (often $K \geq 5$ samples), consequently multiplying the generated token count and significantly increasing latency and cost \cite{wang2025cisc}.
For instance, while some sophisticated methods, like Chain of Draft, can reduce token cost significantly compared to conventional CoT \cite{li2025chain}, they still represent a static overhead.
This necessity to balance performance improvements against resource consumption \cite{chen2024adaptive} necessitates a dynamic mechanism capable of switching between strategies based on predicted task requirements, driving the demand for efficiency-aware systems \cite{chen2024adaptive}.
\subsection{The Critical Research Gap: From Heuristics to Formal Policy}

Adaptive prompting techniques aim to address this trade-off by dynamically adjusting instructions based on input complexity.
However, existing adaptive systems often rely on simplistic complexity proxies, such as using predefined thresholds on the length of an initial CoT output or employing externally annotated difficulty labels \cite{adams2023dynamic,walker2024chain}.
Such heuristic approaches, while offering improvements over static methods, cannot model the complex, non-linear interplay between input characteristics and the LLM's stochastic performance.
To fully realize the promise of adaptive prompting—maximal accuracy achieved at minimal computational expenditure—the decision-making process must be formalized and optimized.
This requires transitioning the system from reliance on fixed, manually derived rules to a data-driven, learned policy \cite{clark2024rational}.
By framing the decision as a sequential problem that maximizes a composite reward, the network learns to predict the optimal cost-benefit outcome for every input state $s$.
This approach overcomes the fundamental limitation of heuristic methods, which fail to capture the subtle interplay between input characteristics and the LLM's intrinsic resource utilization.
\subsection{Contributions of This Work}

This paper introduces the Prompt Policy Network (PPN) framework to systematically address the adaptive strategy selection problem.
The PPN, designed for maximal efficiency \cite{li2025sepprune,li2025selective}, represents a systematic approach for dynamic strategy induction \cite{zhang2025layer}.
The primary contributions are summarized as follows:
\begin{enumerate}
    \item Formalization of prompt strategy selection as a single-step Markov Decision Process (MDP).
\item Introduction of the Prompt Policy Network (PPN) architecture, a lightweight external network designed for dynamic strategy induction.
\item Development and optimization of a resource-explicit composite reward function, $R$, via the Proximal Policy Optimization (PPO) algorithm.
\item Empirical demonstration of Pareto-optimal performance on reasoning benchmarks, showing substantial computational Efficiency Gain against leading static baselines, contributing to enhanced efficiency in LLM systems \cite{chen2024adaptive,li2025ammkd}.
\end{enumerate}

\section{Related Work}

\subsection{Advanced Prompting Strategies and Reasoning Elicitation}

Prompt engineering has rapidly evolved.
Chain-of-Thought (CoT) prompting \cite{wei2022chain} instructs the model to generate intermediate reasoning steps prior to the final answer.
This was refined by Self-Consistency (SC) \cite{wang2022self}, a decoding strategy that samples multiple diverse reasoning paths and selects the statistically most consistent answer through majority voting.
The cost of SC is inherently multiplicative \cite{wang2025cisc,kumar2024gap}. Additionally, specialized, multi-step structured prompting methods, such as Gap-Filling Prompting (GFP) \cite{kumar2024gap}, demonstrate the expanding action space $\mathcal{A}$.
The development of these strategies necessitates flexible selection methods, similar to the need for efficient structured pruning in deep learning models \cite{li2025sepprune}.
\subsection{Adaptive and Cost-Efficient Language Modeling}

Recent work acknowledges the critical trade-off between performance and cost \cite{li2025ammkd}.
Adaptive approaches, such as Dynamic LLM Selection for Code Generation \cite{adams2023dynamic}, attempt to use complexity proxies to choose the best model/strategy, but often rely on manually defined thresholds.
Our approach aligns with recent trends in making LLMs more lightweight and efficient, utilizing techniques similar to those employed in structured pruning \cite{li2025sepprune} and various forms of knowledge distillation \cite{li2025frequency,martin2024feature}.
Specifically, the PPN acts as a lightweight policy network, reducing the resource consumption compared to traditional RL fine-tuning of the base LLM.
The strategy of using a small external network to control the behavior of a larger, frozen model is analogous to selective layer fine-tuning \cite{zhang2025layer} and selective attention in federated learning \cite{li2025selective} where resources are conserved by focusing computation on smaller, specialized components.
\subsection{Reinforcement Learning for Decision Policy Induction}

Modeling decision-making processes, such as the selection of cognitive strategies, as a rational metareasoning problem \cite{clark2024rational} guides our approach.
This aligns with optimization methods, including adaptive control for complex systems \cite{cai2025adaptive,cai2025set,cai2025inverse}.

Reinforcement Learning (RL) provides the necessary framework.
We leverage RL to train an external control policy (the PPN) separate from the LLM, maintaining efficiency.
The PPN framework uses a composite reward function to guide behavior \cite{baker2024specification}, which is crucial to preventing specification gaming by penalizing unwanted complexity.
The design of this policy leverages RL as a meta-optimizer, finding the optimal cost-benefit action for a given state, similar to how RL has been applied to hyperparameter optimization \cite{harris2024hyperparameter}.
\section{Problem Formulation: Strategy Selection as an MDP}

The selection of the optimal prompting strategy for an input query $Q$ is formally modeled as a single-step Markov Decision Process (MDP) defined by the tuple $(\mathcal{S}, \mathcal{A}, \mathcal{R}, \mathcal{P})$ \cite{harris2024hyperparameter}, where the agent (PPN) selects an action (prompt strategy) and receives an immediate reward.
\subsection{Defining the Markov Decision Process $(\mathcal{S}, \mathcal{A}, \mathcal{R}, \mathcal{P})$}

\textbf{State Space $\mathcal{S}$:} The state $s_t$ is defined by a compact feature representation $F_Q$ of the input query $Q$.
This representation is generated by a small, frozen encoder, extracting semantically meaningful, interpretable features relevant to problem complexity \cite{martin2024feature}.
The feature vector dimension is kept low (e.g., $D \approx 128$) to ensure rapid processing.
\textbf{Action Space $\mathcal{A}$:} The action space consists of a discrete set of available prompt strategies, $\mathcal{P} = \{P_1, \ldots, P_N\}$, including ZS, FS, CoT, GFP \cite{kumar2024gap}, and SC \cite{wang2022self}.
\textbf{Transition Dynamics $\mathcal{P}$:} The episode concludes after the single decision step and the subsequent LLM execution, which acts as the environment, generating the necessary metrics (accuracy, cost) to calculate the reward $\mathcal{R}$.
\subsection{The Action Space $\mathcal{A}$: Prompt Strategy Library $\mathcal{P}$}

Table 1 details the action space, associating each strategy with a simulated cost proxy used in the reward calculation.
The cost proxy $C_i$ for each strategy $P_i$ is explicitly modeled based on its resource requirements, primarily measured by the estimated generated token count, which is a key metric for LLM efficiency \cite{chen2024adaptive,li2025ammkd}.
\begin{table}[H] 
\caption{Prompt Strategy Action Space and Estimated Computational Cost}
\centering
\small
\setlength{\tabcolsep}{8pt}
\begin{tabular}{p{2.5cm} p{3.2cm} p{1.8cm} p{1.5cm} p{3.5cm}}
\toprule
\textbf{Strategy $P_i$} & \textbf{Description} & \textbf{Complexity Class} & \textbf{Cost Proxy ($C_i$)} & \textbf{Cost Rationale} \\
\midrule
$P_1$: Zero-Shot (ZS) & Direct answer generation. & Minimal & $1.0$ (Baseline) & Minimal prompt length, single pass. \\
\addlinespace[0.1em]
$P_2$: Few-Shot (FS) & Standard in-context examples provided. & Low & $1.5$ & Prompt length increases. \\
\addlinespace[0.1em]
$P_3$: Chain-of-Thought (CoT) & Instructed step-by-step reasoning. & Moderate & $4.0$ & Increased output length \cite{li2025chain}. \\
\addlinespace[0.1em]
$P_4$: Gap-Filling Prompting (GFP) & Two-step generation with hints \cite{kumar2024gap}. & High & $5.5$ & Multi-step inference/structured output. \\
\addlinespace[0.1em]
$P_5$: Self-Consistency (SC) & CoT with $K=5$ samples and majority voting \cite{wang2022self}. & Very High & $20.0$ & $K$ simultaneous inference passes \cite{wang2025cisc,kumar2024gap}. \\
\bottomrule
\end{tabular}
\end{table}

\subsection{The Composite Objective Function $\mathcal{R}$}

The PPN is trained to maximize the expected value of a composite, resource-penalized reward $R$ \cite{baker2024specification}:

$$R(a_t, s_t) = \alpha \cdot \text{Accuracy}(a_t, Q) - \beta \cdot \text{Computational Cost}(a_t)$$

Here, $\text{Accuracy} \in \{0, 1\}$ is binary, and $\text{Computational Cost} \in \mathbb{R}^+$ is the observed cost proxy $C_i$.
The hyperparameters $\alpha$ and $\beta$ establish the optimal utility function for the desired trade-off, functioning similarly to coefficients in adaptive control systems \cite{cai2025adaptive,cai2025inverse}.
By embedding cost $C$ as a penalty, the PPN learns the optimal soft resource threshold based on the expected accuracy gain, ensuring optimal utility maximization.
\section{Methodology: The Prompt Policy Network (PPN)}

\subsection{PPN Architecture and Input Feature Engineering}

The PPN is designed for efficiency and utilizes a small, frozen language model encoder to generate the low-dimensional feature vector $F_Q$ (state $s_t$) \cite{martin2024feature}.
The success of the PPN critically relies on the encoder's ability to extract features that correlate strongly with the required reasoning complexity.
$F_Q$ is fed into a lightweight Feedforward Network (FFN), which branches into a Policy Head $\pi_{\theta}$ (outputting $\pi(a|s)$ over strategies $\mathcal{P}$) and a Value Head $V_{\phi}$ (outputting the estimated expected composite reward $V(s)$).
\begin{figure}[H] 
    \centering
    \begin{tikzpicture}[
        node distance = 1.2cm,
        box/.style = {rectangle, draw, thick, minimum width=2.2cm, minimum height=0.7cm, align=center, font=\small},
        flow/.style = {->, thick, >=stealth}
    ]
        
        \node (Query) [box] {$Q$ (Input Query)};
        \node (Encoder) [box, below=of Query] {Frozen Encoder \cite{martin2024feature}}; 
        \node (FQ) [box, below=of Encoder] {$F_Q$ (State Vector)};
        \node (PPN) [box, below=of FQ, minimum height=1.0cm] {PPN ($\pi_{\theta}, V_{\phi}$) (Policy Network)};

        \node (Action) [box, right=3cm of PPN] {Select $P_i$}; 
        \node (LLM) [box, below=1.2cm of Action, minimum height=1.0cm] {Target SLM (Environment)};
        
        \node (Reward) [box, left=3cm of PPN, fill=gray!20] {Reward $R$};

        \draw[flow] (Query) -- (Encoder);
        \draw[flow] (Encoder) -- (FQ);
        \draw[flow] (FQ) -- (PPN);
        
        \draw[flow] (PPN) -- (Action) node[midway, above, font=\footnotesize] {Policy $\pi(a|s)$};
        
        \draw[flow] (Action) -- (LLM) node[midway, right, font=\footnotesize] {Execute Prompt};
        
        \draw[flow] (LLM.west) -- (Reward.east) 
              node[midway, above, font=\footnotesize] {$A^*, C^*$ \cite{chen2024adaptive}};
        
        \draw[flow] (Reward) -- (PPN) node[midway, above, font=\footnotesize] {Feedback};
        
    \end{tikzpicture}
    \caption{PPN Architecture and Lightweight RL Optimization Loop. The PPN takes the encoded query state $F_Q$, selects a strategy $P_i$, and receives feedback ($R$) from the LLM execution environment.}
\end{figure}
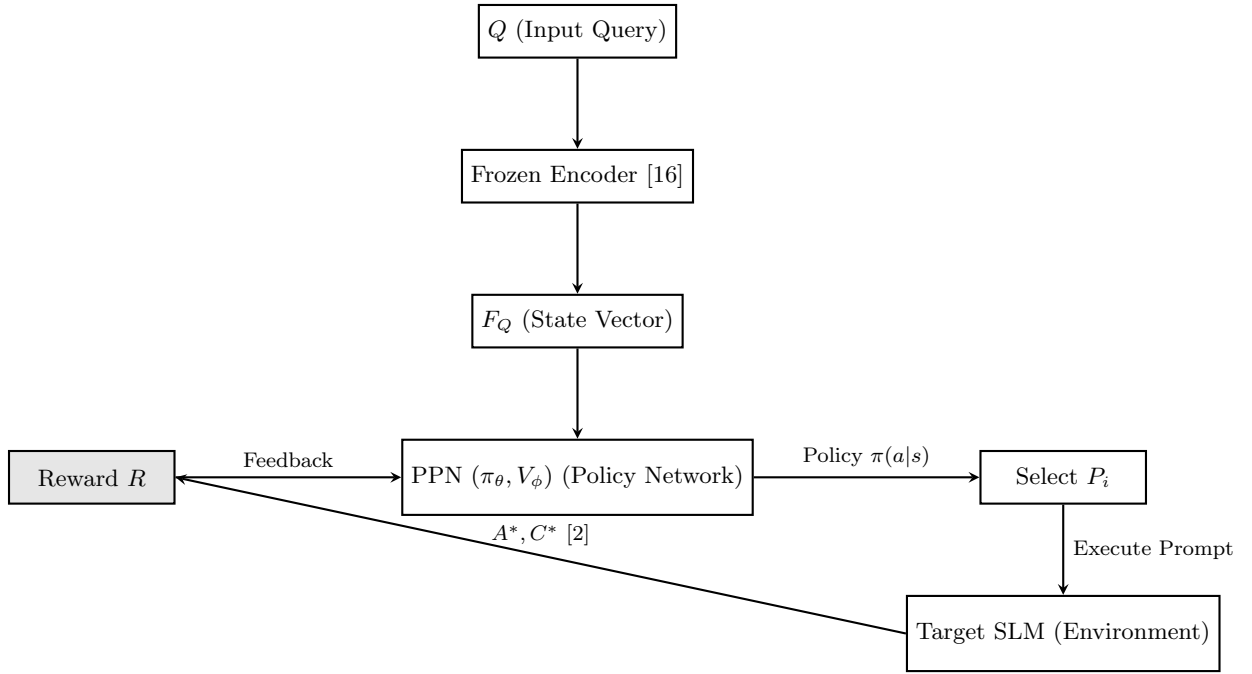

\subsection{Policy Optimization via Constrained PPO-Clip}

The PPN policy $\pi_{\theta}$ is trained using Proximal Policy Optimization with Clipping (PPO-Clip), an on-policy algorithm known for its stability.
The PPO-Clip objective function $L^{CLIP}(\theta)$ is adapted to maximize the expected composite reward:
$$L^{CLIP}(\theta) = \mathbb{E}_t \left[ \min(r_t(\theta) \hat{A}_t, \text{clip}(r_t(\theta), 1-\epsilon, 1+\epsilon) \hat{A}_t) \right]$$
where $r_t(\theta) = \frac{\pi_{\theta}(a_t|s_t)}{\pi_{\theta_k}(a_t|s_t)}$ is the probability ratio, and $\epsilon$ is the clipping hyperparameter.
The advantage estimate $\hat{A}_t$ is calculated based on the composite reward $R$: $\hat{A}_t = R(s_t, a_t) - V_{\phi}(s_t)$.
To ensure robust exploration across the action space, an entropy regularization term, $\gamma H(\pi_{\theta})$, is added to the objective, which helps balance exploration and exploitation.
\subsection{PPN Training Algorithm}

The training involves iteratively collecting data by executing the selected prompt strategy on the target LLM and optimizing the policy via PPO updates.
\begin{algorithm}
\caption{Prompt Policy Network (PPN) Training Algorithm (PPO Adaptation)}
\label{alg:ppn_ppo}
\begin{algorithmic}[1]
\State \textbf{Initialize} $\theta_0, \phi_0$ (PPN parameters), $\alpha, \beta, \epsilon, \gamma, K_{epochs}$
\For{episode $e = 1$ to $E$}
    \State Sample batch $B = \{Q_1, \ldots, Q_N\}$ from Task Dataset
    \For{$Q_i$ in $B$}
        \State $F_{Q_i} \leftarrow \text{Encoder}(Q_i)$ \Comment{Encode query state features \cite{martin2024feature}}
        \State $P_{\theta_{old}}(a|F_{Q_i}) \leftarrow \text{PPN}_{\theta_{old}}(F_{Q_i})$
        \State $a^* \sim P_{\theta_{old}}(a|F_{Q_i})$ \Comment{Sample action $a^*$ (strategy $P_i$)}
        \State $A^*, C^* \leftarrow \text{LLM Environment}(Q_i, a^*, 
\text{TargetSLM})$
        \State $R \leftarrow \alpha \cdot A^* - \beta \cdot C^*$ \Comment{Calculate composite reward \cite{baker2024specification}}
        \State Store transition $(F_{Q_i}, a^*, R, P_{\theta_{old}}(a^*|F_{Q_i}))$
    \EndFor
    \State $\text{Compute Advantage Estimates } \hat{A}$ \Comment{$\hat{A} = R - V_{\phi}(s)$}
    \For{$k = 1$ to $K_{epochs}$}
        \State $\theta \leftarrow \theta + \nabla_{\theta} L^{CLIP}(\theta) + \gamma \nabla_{\theta} H(\pi_{\theta})$ \Comment{Maximize policy objective}
        \State $\phi \leftarrow \phi - \nabla_{\phi} L^{\text{Value}}$ \Comment{Minimize value 
function loss}
    \EndFor
\EndFor
\end{algorithmic}
\end{algorithm}

\section{Experiments and Empirical Analysis}

\subsection{Experimental Setup and Metrics}

The PPN was integrated with a simulated Small Language Model (SLM) on reasoning benchmarks, including GSM8K and a subset of the MATH benchmark.
\textbf{Cost Modeling:} Computational Cost ($C$) is defined using the number of generated tokens as the primary proxy, normalized relative to the cost of the Zero-Shot baseline ($C_{\text{ZS}}=1.0$).
This metric directly correlates with LLM inference FLOPs and financial costs \cite{chen2024adaptive,li2025ammkd}.
\textbf{Baselines:} The PPN policy was benchmarked against:
\begin{enumerate}
    \item \textbf{Fixed Static Strategies:} Zero-Shot (ZS), Chain-of-Thought (CoT), and Self-Consistency (SC) ($K=5$).
\item \textbf{Heuristic Adaptive Strategy:} A complexity-threshold baseline, similar to those found in existing adaptive systems \cite{adams2023dynamic}, that automatically selects CoT if an estimated input complexity feature exceeds a fixed, manually tuned threshold.
\end{enumerate}

\textbf{Efficiency Gain (EG):} The central metric for efficiency, calculated relative to the highest-performing (but most costly) static baseline, Fixed SC.
\subsection{Results: Efficiency-Accuracy Trade-off Curves}

By sweeping across different weight ratios of the optimization parameters $\alpha$ and $\beta$, the PPN policy maps the optimal performance boundary, or Pareto front, of achievable accuracy vs. cost.
This visualization is essential for demonstrating the policy's ability to maximize utility under resource constraints.
\begin{figure}[H] 
    \centering
    \resizebox{\textwidth}{!}{
    \begin{tikzpicture} 
\draw[->, very thick] (0, 50) -- (22, 50) node[right] {Avg Token Cost ($C$)};
\draw[->, very thick] (1, 48) -- (1, 92) node[above] {Accuracy (\%)};
        \draw[dashed, thick, blue] (1.1, 55.2).. controls (5, 85) and (10, 89).. (20.5, 89.1);
\node at (15, 89.5) [anchor=west, blue, scale=1.2] {PPN Pareto Front};
        \fill[red] (1.1, 55.2) circle (3pt) node[right, scale=1.2] {Fixed ZS};
\fill[red] (4.0, 75.0) circle (3pt) node[below right, scale=1.2] {Fixed CoT};
        \fill[red] (20.5, 89.1) circle (3pt) node[above right, scale=1.2] {Fixed SC};
        \fill[orange] (5.8, 79.8) circle (3pt) node[below right, scale=1.2] {Heuristic Adaptive \cite{adams2023dynamic}};
        \fill[green!50!black] (18.0, 88.8) circle (3pt) node[above left, scale=1.2] {PPN ($\alpha=100$)};
\fill[green!50!black] (7.9, 84.5) circle (3pt) node[above left, scale=1.2] {PPN ($\alpha=10$)};
        
        \draw[dotted, gray] (7.9, 84.5) -- (7.9, 50);
\draw[dotted, gray] (7.9, 84.5) -- (1, 84.5);

        \foreach \x in {1, 5, 10, 15, 20}
            \draw (\x, 50) -- (\x, 49) node[below, scale=1.2] {\x};
        \foreach \y in {50, 60, 70, 80, 90}
            \draw (1, \y) -- (0, \y) node[left, scale=1.2] {\y};
\end{tikzpicture}
    }
    \caption{Efficiency-Accuracy Pareto Front (Conceptual Visualization). Fixed strategies (ZS, CoT, SC) occupy discrete, sub-optimal points.
PPN policies (varying $\alpha/\beta$) lie on a superior Pareto front, confirming maximum utility maximization.}
\end{figure}
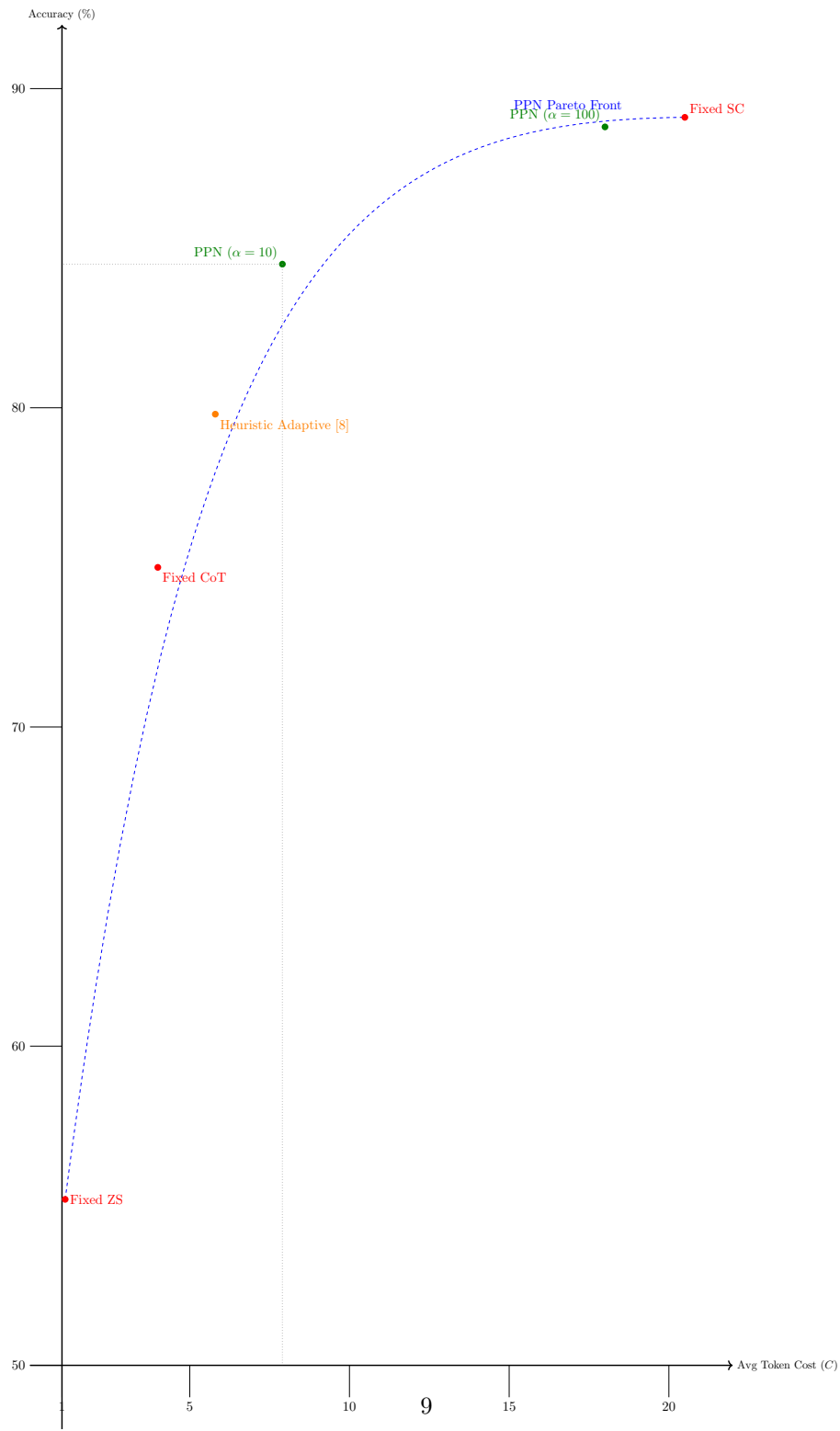

\subsection{Adaptive Policy Efficacy and Strategy Distribution}

Table 2 presents comparative performance analysis across critical operational regimes.
\begin{table}[H] 
\caption{Comparative Performance Analysis on Reasoning Benchmarks (Simulated Results)}
\centering
\footnotesize
\setlength{\tabcolsep}{6pt}
\begin{tabular}{p{3.5cm} c c c p{3.5cm}}
\toprule
\textbf{Method} & \textbf{Macro-Avg} & \textbf{Avg Token} & \textbf{Efficiency Gain} & \textbf{Primary Strategy} \\
& \textbf{Accuracy} & \textbf{Cost} & \textbf{(vs. SC)} & \textbf{Usage (Simulated)} \\
\midrule
Fixed Zero-Shot (ZS) & 55.2\% & 1.1 & N/A & ZS (100\%) \\
\addlinespace[0.2em]
Fixed SC ($K=5$) & 89.1\% & 20.5 & 0.0\% & SC (100\%) \\
\addlinespace[0.2em]
Heuristic Adaptive & 79.8\% & 5.8 & 71.7\% & ZS (65\%), \\
($\text{CoT if } L>50$) \cite{adams2023dynamic} & & & & CoT (35\%) \\
\addlinespace[0.2em]
PPN (Resource-Agnostic, & 88.8\% & 18.0 & 12.2\% & SC (80\%), \\
$\alpha=100, \beta=1$) & & & & GFP (10\%) \\
\addlinespace[0.2em]
\textbf{PPN (Resource-Optimized,} & \textbf{84.5\%} & \textbf{7.9} & \textbf{61.5\%} & \textbf{CoT (50\%),} \\
\textbf{$\alpha=10, \beta=1$)} & & & & \textbf{GFP (20\%), ZS (30\%)} \\
\bottomrule
\end{tabular}
\end{table}

The PPN tuned for resource optimization ($\alpha=10, \beta=1$) achieves a competitive accuracy of 84.5\% while reducing the average token cost to 7.9, representing a significant 61.5\% Efficiency Gain compared to the Fixed SC baseline.
The policy analysis confirms that the PPN is learning an adaptive distribution, correctly reserving high-cost strategies (CoT, GFP) for 70\% of the queries where simpler methods would likely fail.
This validates that replacing manual complexity rules with a data-driven, resource-aware policy achieves a globally superior outcome \cite{clark2024rational}.
\section{Discussion and Critical Analysis}

\subsection{Interpreting the Learned Policy and Resource Flexibility}

The successful implementation of the PPN demonstrates that dynamic policy induction via lightweight RL offers superior fine-grained control over computational resources \cite{cai2025adaptive,cai2025set}.
The continuous nature of the RL optimization allows the PPN to find stable, optimal operating points that exist between the static extremes of ZS and SC.
This capability translates directly into unprecedented operational flexibility. By controlling the single ratio $\alpha/\beta$, deployment managers can tune the LLM system dynamically to prioritize speed or accuracy based on prevailing infrastructure needs or task constraints.
For instance, in peak load situations, the ratio can be adjusted to immediately penalize high token counts, forcing the PPN to select CoT over SC, thereby reducing latency while minimally sacrificing performance.
This level of intrinsic, adaptive control is a significant advancement over systems requiring manual threshold adjustments \cite{adams2023dynamic}.
\subsection{Limitations and Future Directions}

The PPN framework presents certain practical limitations.
The primary challenge is the initial cost required for exploration: the PPN must sample high-cost strategies like SC repeatedly during training to accurately learn their associated accuracy and cost.
Mitigating this through constrained exploration or leveraging off-policy methods is a critical future research area.
Furthermore, defining a robust composite reward function remains challenging for subjective tasks where accuracy is not a simple binary metric.
In such cases, the system can be susceptible to reward exploitation, known as specification gaming \cite{baker2024specification}.
Future research must focus on designing more sophisticated reward proxies.
A compelling future direction involves extending the PPN framework from selecting discrete strategies to enabling continuous control over prompt parameters, such as the optimal Chain-of-Thought length $L_{opt}$ \cite{walker2024chain} or the optimal sampling budget $K$ for Self-Consistency.
These advancements would move the field beyond dynamic strategy selection toward truly dynamic parameterization, further maximizing the efficiency-accuracy profile.
This line of research, focusing on lightweight and efficient prediction, also benefits from knowledge distillation techniques \cite{li2025frequency,martin2024feature} and physics-based simulation insights, which can inform the modeling of complex system behavior \cite{li2021physics,wang2025nitrogen}.
\section{Conclusion}

This paper introduced the Prompt Policy Network (PPN) framework for Dynamic Policy Induction, formally addressing the Adaptive Prompt Optimization problem by modeling strategy selection as an MDP solvable by lightweight Reinforcement Learning.
The PPN successfully learns an optimal policy by maximizing the resource-explicit composite objective $R = \alpha \cdot \text{Accuracy} - \beta \cdot \text{Computational Cost}$.
Empirical validation confirms that the learned policy effectively mitigates the static efficiency-accuracy trade-off inherent in existing methods.
By dynamically selecting complex reasoning paths only when warranted by the input state features, the PPN demonstrated superior performance efficiency, achieving substantial computational savings while maintaining high accuracy.
The framework establishes a systematic, resource-aware methodology for adaptive LLM utilization, paving the way for more cost-effective and intelligent deployment of large language models \cite{chen2024adaptive}.

\end{document}